\documentclass[10pt,twocolumn,letterpaper]{article}

\usepackage[pagenumbers]{cvpr} % To force page numbers, e.g. for an arXiv version

\usepackage[dvipsnames]{xcolor}

\usepackage{xspace} % for easy naming
\usepackage{bm} % for bold gamma

\def\mL{{\mathcal L}}

\DeclareMathAlphabet\mathbfcal{OMS}{cmsy}{b}{n}
\def\0{{\bf 0}}
\def\1{{\bf 1}}

\def\bE{{\bf E}}

\def\bI{{\bf I}}

\def\bU{{\bf U}}
\def\bV{{\bf V}}

\def\bX{{\bf X}}

\def\bZ{{\bf{Z}}}

\def\be{{\bf e}}

\def\bu{{\bf u}}
\def\bv{{\bf v}}

\def\bz{{\bf z}}

\def\mmR{{\mathbb R}}

\def\bX{{\bf X}}

\def\bz{{\bf z}}

\def\ie{\emph{i.e.,~}} 
 
 \def\vs{\emph{vs.~}}
\def\wrt{{w.r.t.~}}

\newcommand{\bestimprove}[2]{\textbf{#1}$_\text{\textcolor{teal}{~(+#2)}}$}

\usepackage{pifont} % for √ and ×

\newcommand\invisiblesection[1]{\noindent\textbf{#1}.} % SMIL def

\let\AtPageUpperLeft\undefined

\let\AtTextUpperLeft\undefined

\let\AddToShipoutPicture\undefined

\let\gridSetup\undefined

\makeatletter
\let\ESO@isMEMOIR\undefined
\let\ESO@HookI\undefined
\let\ESO@HookII\undefined
\let\ESO@HookIII\undefined
\let\ESO@gridunitname\undefined
\let\ESO@gridunit\undefined
\let\ESO@griddelta\undefined
\let\ESO@griddeltaY\undefined
\let\ESO@gridDeltaY\undefined
\let\ESO@gridcolor\undefined
\let\ESO@subgridcolor\undefined
\let\ESO@subgridstyle\undefined
\let\ESO@gap\undefined
\let\ESO@yoffsetI\undefined
\let\ESO@yoffsetII\undefined
\let\ESO@gridlines\undefined
\let\ESO@subgridlines\undefined
\let\ESO@hline\undefined
\let\ESO@vline\undefined
\let\ESO@Hline\undefined
\let\ESO@Vline\undefined
\let\ESO@fcolorbox\undefined
\let\ESO@color\undefined
\let\ESO@colorbox\undefined
\let\ESO@div\undefined
\let\ESO@gridpicture\undefined
\let\ESO@labelfactor\undefined
\let\ESO@gridDelta\undefined
\makeatother

\definecolor{cvprblue}{rgb}{0.21,0.49,0.74}
\usepackage[pagebackref,breaklinks,colorlinks,citecolor=cvprblue]{hyperref}

\usepackage{booktabs}
\usepackage{multirow}
\usepackage{bbding}
\usepackage{xcolor}
\usepackage{pdfpages} % for supplementary

\def\shortname{CG-VLM\xspace}
\def\name{Contrastive and Generative Aligned VLM\xspace}

\definecolor{favorable}{RGB}{157,195,230}
\definecolor{unfavorable}{RGB}{247,197,159}

\def\paperID{1400} % *** Enter the Paper ID here
\def\confName{CVPR}
\def\confYear{2024}

\title{Contrastive Vision-Language Alignment Makes Efficient Instruction Learner}

\author{Lizhao Liu$^{1,2}$\footnotemark[1] ~~Xinyu Sun$^{1}$\footnotemark[1] ~~Tianhang Xiang$^1$~~Zhuangwei Zhuang$^1$ ~~Liuren Yin$^{3}$~~~Mingkui Tan$^{1,2}$\footnotemark[2]\\
$^1$South China University of Technology~~$^2$PengCheng Laboratory~~$^3$Duke University \\
{\tt\small \{selizhaoliu, csxinyusun, sexiangtianhang, z.zhuangwei\}@mail.scut.edu.cn,} \\ {\tt\small liuren.yin@duke.edu, mingkuitan@scut.edu.cn} \\
{\small Code is available at: \url{https://github.com/lizhaoliu-Lec/CG-VLM}}
}

\begin{document}
\maketitle

\renewcommand{\thefootnote}{\fnsymbol{footnote}}
\footnotetext[1]{Equal contribution.}
\footnotetext[2]{Corresponding author.}
\renewcommand{\thefootnote}{\arabic{footnote}}

\begin{abstract}
% We study the task of extending the multimodal capability of large language models (LLMs) by vision-language learning.
We study the task of extending the large language model (LLM) into a vision-language instruction-following model. This task is crucial but challenging since the LLM is trained on text modality only, making it hard to effectively digest the visual modality.
% We study the task of learning instruction-following vision-language models (VLM) from pre-trained vision transformers (ViT) and large language models (LLM). 
To address this, existing methods typically train a visual adapter to align the representation between a pre-trained vision transformer (ViT) and the LLM by a generative image captioning loss.
% To address this, existing methods typically connect the ViT and the LLM with a visual adapter. Since the semantic spaces of different pre-trained models are naturally not matched, given the visual features transformed by the visual adapter as the prefix, a vision-language alignment step is conducted to learn the visual adapter by a generative image captioning loss. 
However, we find that the generative objective can only produce weak alignment for vision and language, making the aligned vision-language model very hungry for the instruction fine-tuning data. In this paper, we propose \shortname that applies both \textbf{\underline{C}}ontrastive and \textbf{\underline{G}}enerative alignment objectives to effectively align the representation of \textbf{\underline{V}}iT and L\textbf{\underline{LM}}. Different from image level and sentence level alignment in common contrastive learning settings, \shortname aligns the image-patch level features and text-token level embeddings, which, however, is very hard to achieve as no explicit grounding patch-token relation provided in standard image captioning datasets. To address this issue, we propose to maximize the averaged similarity between pooled image-patch features and text-token embeddings. Extensive experiments demonstrate that the proposed \shortname produces strong vision-language alignment and is an efficient instruction learner. For example, using only 10\% instruction tuning data, we reach 95\% performance of state-of-the-art method LLaVA~\cite{liu2023improved} on the zero-shot ScienceQA-Image benchmark. 
% First, to overcome the semantic spaces misalignment rooted in LLM and ViT, we apply contrastive loss on linear-transformed visual features and prompted language representation. Compared to existing generative alignment, \shortname is much more efficient by learning only a linear layer and a \texttt{[CLS]} prompt in a contrastive way. Second, to address the semantic granularity mismatch issue, we propose to linear probe the aligned representation and select the combination with the best classification accuracy. Extensive experiments demonstrates that the proposed \shortname reach superior performance than existing methods in a much more efficient way.
\end{abstract}
\vspace{-20pt}
\section{Introduction}
\label{sec:intro}

\begin{figure}[!t]
    \centering
    \includegraphics[width=1.0\linewidth]{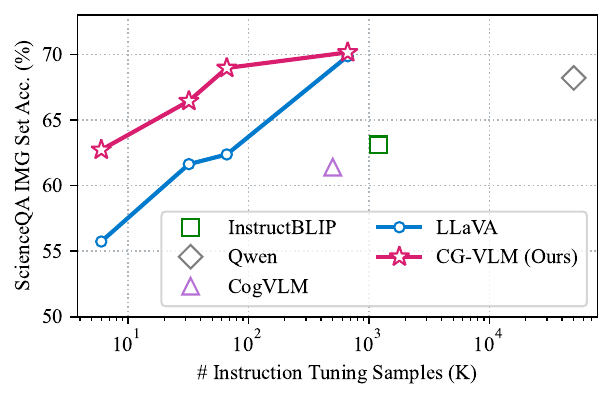}
    \vspace{-20pt}
    \caption{Comparisons to state-of-the-art visual instruction methods on zero-shot ScienceQA~\cite{lu2022sqa} image set \wrt~different amount of instruction tuning data. 
    }
    \label{fig:fig1}
    \vspace{-20pt}
\end{figure}

Visual instruction-following model has been considered as the bedrock for the general-purpose assistant~\cite{li2023blip,dai2023instructblip} due to its ability to understand vision-language information and follow language instruction, drawing a large amount of attention from researchers~\cite{li2023blip,dai2023instructblip,gao2023llama,zhang2023llama,zhu2023minigpt,liu2023visual,liu2023improved,lin2023learning,chen2023a2nav}. Early attempts at vision instruction following models are restricted to image captioning~\cite{vinyals2016show,johnson2016densecap} and visual question answering~\cite{li2019visualbert,lu2019vilbert} tasks, limiting their application scenarios such as open-ended vision-language assistant~\cite{zhang2023llama}. Recently, attracted by the appealing instruction-following ability of large language models (LLM)~\cite{ouyang2022training,touvron2023llama,chiang2023vicuna}, researchers turn to adapt the LLM into a vision-language instruction following models, with the help of pre-trained vision transformer (ViT)~\cite{dosovitskiy2020image} from CLIP~\cite{alec2021clip}. The main focus is on how to combine the pre-trained ViT and LLM in order to complete the complex visual instruction task. 

To this end, existing methods often require a vision-language alignment step or mechanism that aligns the vision and text modalities by learning a visual adapter. There are mainly two lines of work: heavyweight adaptation~\cite{li2023blip,zhu2023minigpt,dai2023instructblip,alayrac2022flamingo,bai2023qwen} and lightweight adaptation~\cite{gao2023llama,zhang2023llama,liu2023visual,liu2023improved}. BLIP-2~\cite{li2023blip}, MiniGPT4~\cite{zhu2023minigpt} and InstructBLIP~\cite{dai2023instructblip} leverage a heavy visual adapter Q-former, a BERT~\cite{kenton2019bert} model, to align different modalities. Flamingo~\cite{alayrac2022flamingo} propose to use perceiver~\cite{jaegle2021perceiver} and gated xattn-dense layers to respectively extract and inject vision-language features into a pre-trained Chinchilla LLM~\cite{hoffmann2022training}. Different from them, lightweight adaptation methods such as LLaMA-Adapter~\cite{gao2023llama,zhang2023llama} and LLaVA~\cite{liu2023visual,liu2023improved} propose to learn a linear or MLP projector that connects the pre-trained ViT and LLM. During the vision-language alignment step, the visual features from ViT are projected to the embedding space of LLM by the visual adapter. Then, the projected visual features are fed into LLM and are optimized by the image captioning loss in a generative (\ie auto-regressive) way.

% \begin{table}[!t]
%     \centering
%     \resizebox{0.8\linewidth}{!}{\small
%         \begin{tabular}{l|cr}
%         % \toprule
%         \hline
%         Method & Align Type & Top-1 Acc. (\%) \\
%         \hline\hline
%         \textcolor{gray}{CLIP}~\cite{alec2021clip} & \textcolor{gray}{Con} & 
%         \textcolor{gray}{69.03} \\
%         \hline
%         LLaVA~\cite{liu2023improved} & Gen & 0.14 \\
%         % CTC (Ours) & Con & 21.74 \\
%         CTC (Ours) & Con + Gen & 13.61 \\
%         % \without & \without & 44.2 & 61.8 & 52.9 & 64.9 \\
%         % \hline
%         % \with & \without & \noimprove{41.6}{2.6} & \noimprove{58.6}{3.2} & \noimprove{51.1}{1.8} & \noimprove{63.6}{1.3} \\
%         % \with & \with & \bestimprove{52.2}{6.7} & \bestimprove{63.8}{2.0} & \bestimprove{59.3}{6.4} & \bestimprove{66.3}{1.4}\\
%         % \bottomrule
%         \hline
%         \end{tabular}
% }
% \caption{Zero-shot evaluation on ImageNet using the text prompt template \texttt{A photo of a \{class name\}}. Con and Gen are short for Contrastive and Generative, respectively.}
% \label{tab:imgnet_cls}
% \end{table}

\begin{figure*}[!t]
    \centering
    \includegraphics[width=1.0\linewidth]{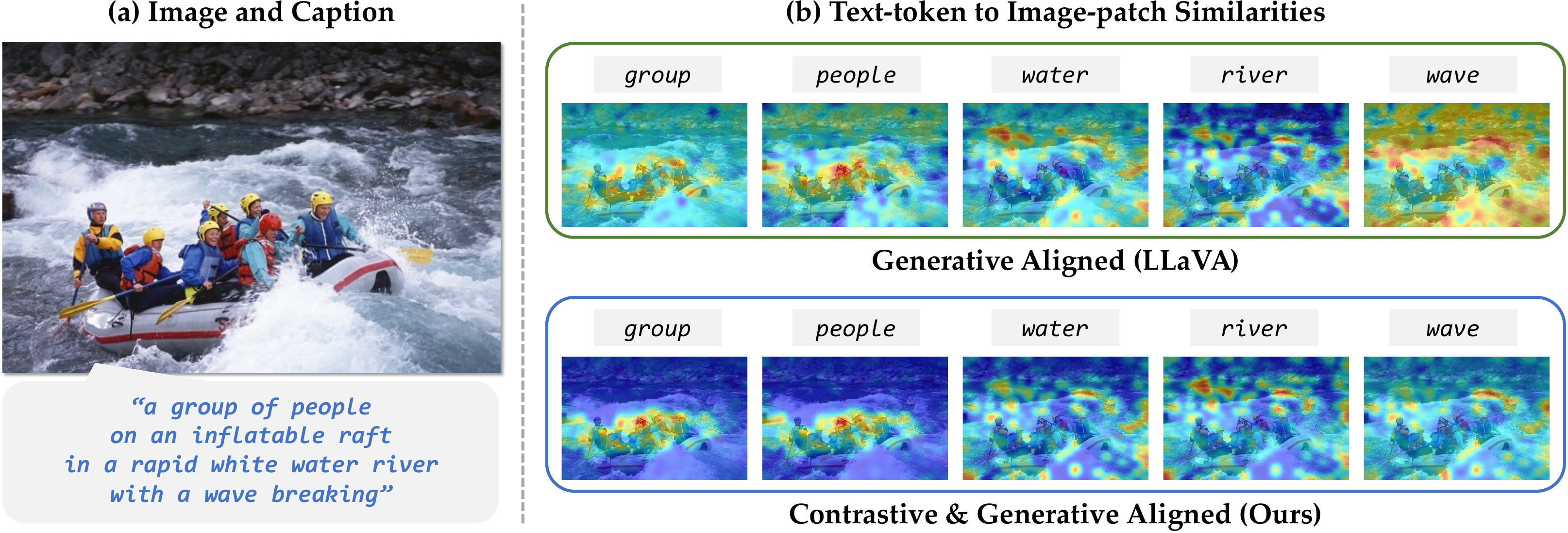}
    \caption{Comparisons of the image features projected by different visual adapters. With generative objective only, the projected visual features of LLaVA~\cite{liu2023improved} have large similarities to all visual concepts. In contrast, with the proposed contrastive and generative aligned visual adapters, our \shortname provides \textit{only} large similarities to the corresponding concepts.\protect\footnotemark
    }
    \label{fig:motivation}
    % \vspace{-10 pt}
\end{figure*}

However, we suspect that training the visual adapter in a generative manner may not align vision-language modalities effectively. To verify this, we visualize the cosine similarities between the image patch and text token, and the results are shown in Figure~\ref{fig:motivation}. We find that the text token features often have a large cosine similarity to a majority of vision patch features, showing the weak alignment between vision-language modalities. Moreover, this weak alignment makes the model very data-hungry for the vision instruction fine-tuning data. From Figure~\ref{fig:fig1}, with a generative aligned visual adapter, the accuracy of state-of-the-art method LLaVA~\cite{liu2023improved} on Science-QA~\cite{lu2022sqa} image set drops significantly as the amount of instruction tuning samples decreases. On one hand, instruction tuning data is very crucial for generalization to vision-language tasks~\cite{dai2023instructblip}. On the other, high-quality instruction data is very hard to collect since the process is time-consuming and less well-defined when human crowd-scouring is considered~\cite{liu2023visual}. In this sense, how to align the vision language effectively remains an important but unsolved question.  
% \lec{Emphasize the importance of instruction data efficiency. Collecting high-quality instruction data is very difficult.}

% we use ImageNet~\cite{deng2009imagenet} zero-shot probing accuracy like CLIP~\cite{alec2021clip} to evaluate the vision-language alignment quality. As shown in Table~\ref{tab:imgnet_cls}, the zero-shot probing accuracy of the generatively-trained visual adapter is near random guesses. Moreover, as depicted in Figure~\ref{fig:motivation}, the cosine similarity map between vision patch features and text token features also shows weak alignment between vision-language modalities. Last but not least, this weak alignment makes the model very data-hungry for the vision instruction fine-tuning data. As shown in Figure~\ref{fig:fig1}, with a generative-only aligned visual adapter, the accuracy of Science-QA~\cite{lu2022sqa} image-related questions drops significantly.

In this paper, we propose \name (\shortname) to effectively align the vision-language modalities from pre-trained ViT and LLM. To be specific, we use contrastive loss to align the vision and text features from the visual adapter and LLM embedding layer, respectively. 
However, as the averaged token embeddings of LLM layers are very ambiguous among different sentences~\cite{ethayarajh2019contextual}, 
% However, 
how to align the visual features to a sequence of token embeddings from the caption remains an unknown question. To this end, we propose to average the similarities between pooled image features and token features from one sentence as the image-sentence similarities. Then, we optimize the image-text similarities from batch data like CLIP~\cite{alec2021clip}. Though the proposed contrastive objective provides better vision-language alignment performance, the average mechanism of image-sentence alignment inevitably focuses on the dominant visual concepts, resulting in sub-optimal vision-language alignment of the nondominant concepts or semantics. To address this issue, we preserve the original generative objective that should attend to the detailed and rich visual concept in order to complete the dense image captioning task. With our proposed \shortname, we make better visual instruction tuning in two aspects: \textbf{1)} a visual instruction tuning data efficient vision-language model that achieves most of the performance even trained with only 10\% amount of the original data. \textbf{2)} preserves more text ability from the original LLM and achieves better results on the text-only instruction or question. We summarize our main contributions as follows:
\begin{itemize}
    \item To the best of our knowledge, we are the first to analyze the vision-language representation alignment issue in the visual instruction tuning task and find that existing works can not effectively align vision-language with only the generative objective given limited instruction tuning data. 

    \item To effectively align vision-language modalities, we propose to use the contrastive objective on the image-sentence similarities. Moreover, to remedy the sub-optimal alignment of non-dominant concepts, we also preserve the original generative objectives to focus more on the rich and detailed semantics.

    \item Extensive experiments on popular visual question answering and human instruction following benchmarks show that the proposed \shortname surpasses the baseline method by a large margin under the instruction efficient settings and achieves competitive results compared to methods that use 100$\times$ instruction tuning data. 
\end{itemize}
\section{Related Works}
\label{sec:related_works}

\noindent\textbf{Vision-Language Models} (VLM) aims to perceive and understand vision-language modalities, in order to provide text response to the corresponding query. Early works focus on image captioning~\cite{johnson2016densecap,vinyals2016show} and visual question answering~\cite{li2019visualbert,lu2019vilbert}, which are restricted to specific tasks or domains. With the great progress made in vision or language models pre-training~\cite{kenton2019bert,radford2019language,he2020momentum,chenempirical}, researchers dive into vision-language pre-training~\cite{li2019visualbert,chen2020uniter,li2020unicoder,yu2022coca} to broaden the application boundary of VLM. VisualBERT~\cite{li2019visualbert} treats image patch embedding as text token embedding and applies mask language loss and sentence-image prediction loss to pre-train a BERT-like model. UNITER~\cite{chen2020uniter} and Unicoder-VL~\cite{li2020unicoder} extract the visual features by faster R-CNN~\cite{ren2015faster} and further propose masked region modeling or classification objective to improve visual understanding. CoCa~\cite{yu2022coca} combines both the merits of contrastive loss and image captioning loss to pre-train the VLM for both visual retrieval and image captioning tasks. However, without very large-scale pre-train data and intensive computation cost, the language ability of these pre-trained VLMs typically falls short for the large language models~\cite{dai2023instructblip,wang2023large}.

\noindent\textbf{Instruction-Following Models} are developed to improve the instruction-following ability~\cite{christiano2017deep} of large language models (LLM) that have shown strong understanding and reasoning ability since the birth of GPT-3~\cite{brown2020language}. Pioneer works such as InstructGPT~\cite{ouyang2022training} and AnthropicLLM~\cite{bai2022training} show that fine-tuning LLM with instruction datasets produces LLM with more natural and accurate responses. Recently, to improve the performance of open-source LLMs, LLaMA~\cite{touvron2023llama} propose to train the LLM with more tokens given various inference budgets. Subsequently, the instruction-following ability of LLaMA is further improved by Alpaca~\cite{taori2023stanford}, Vicuna~\cite{chiang2023vicuna} and LLaMA 2~\cite{touvron2023llama2}. With great progress made in the language modality, how to endow the instruction-following ability to vision-language modalities like GPT-4~\cite{openai2023gpt4} remains an open challenge.

\noindent\textbf{Visual Instruction Models} aims to incorporate the vision-language understanding and reasoning ability into pre-trained unimodal LLMs given pre-trained vision models such as vision transformers (ViT)~\cite{dosovitskiy2020image}. 
% Existing methods often consist of two stages: vision-language alignment and visual instruction tuning. 
The key to transforming unimodal LLM and ViT into a vision-language model lies in the vision-language alignment stage. In this stage, visual adapters are used to connect unimodal models and trained by image captioning data. Considering the scale of visual adapters, there are two lines of adaptation strategies: heavyweight adaptation and lightweight adaptation. Heavyweight adaptation typically aligns the vision language modalities with a deep module~\cite{alayrac2022flamingo,li2023blip,zhu2023minigpt,dai2023instructblip,bai2023qwen} or adds new branches to the LLM~\cite{wang2023cogvlm,alayrac2022flamingo}. BLIP-2~\cite{li2023blip}, MiniGPT4~\cite{zhu2023minigpt} and InstructBLIP~\cite{dai2023instructblip} leverages Q-former, a BERT module, to extract useful visual features for LLM. Qwen~\cite{bai2023qwen} uses a cross-attention module and tunes ViT during alignment. CogVLM~\cite{wang2023cogvlm} adds a different QKV matrix and FFN layer to LLM for visual features. Flamingo~\cite{alayrac2022flamingo} uses perceiver extract visual features, which are incorporated into newly added gated xattn-dense layers of LLM. Different from them, lightweight adaptation methods~\cite{zhang2023llama,gao2023llama,liu2023visual,liu2023improved,chen2023shikra} align the vision language modalities with simply a linear or MLP layer, leaving the ViT untouched. LLaMA Adapter V1~\cite{zhang2023llama} and V2~\cite{gao2023llama} project the visual features by a linear layer and inject learnable soft prompts and the projected visual features for each LLM layer with a zero-init attention mechanism. LLaVA~\cite{liu2023visual}, LLaVA 1.5~\cite{liu2023improved} and Shikra~\cite{chen2023shikra} use a linear layer or an MLP module to transform the visual features that are fed directly into the LLM. Despite the great progress made by existing methods, the vision and language are weakly aligned due to image captioning loss, a generative objective.

% BLIP-2~\cite{li2023blip} xxx. Instruct-BLIP~\cite{dai2023instructblip} xxx.  LLaMA-Adapter~\cite{zhang2023llama} only uses 1.2M learnable parameters upon the frozen LLaMA 7B model and costs less than one hour for fine-tuning on 8 A100 GPUs. LLaMA-Adapter prepends a set of learnable prompts in the last 30 layers of LLaMA. LLaMA-Adapter V2~\cite{gao2023llama} tune the parameter of pretrained LLM and distribute the visual prompts to the shallow layers of LLM to further improve the performance. Since the scale of multimodal alignment data and multimodal instruction tuning data varies by a large margin \eg~10$\times$~\cite{liu2023visual,gao2023llama}, LLaMA Adapter V2 proposes to optimize disjoint group of parameters of LLaMA. MiniGPT-4~\cite{zhu2023minigpt} aligns a frozen visual encoder with a frozen LLM using one projection layer.
\section{\name}
In this section, we first provide the problem definition and motivation of our \shortname in Section~\ref{sec:motivation}. Then, we provide an overview of the proposed \shortname in Section~\ref{sec:overview}. In Section~\ref{sec:pretrain}, we present our contrastive and generative alignment objectives to connect the vision-language modalities. Last, we present our pipeline for visual instruction tuning in Section~\ref{sec:finetune}.

\footnotetext{More visualization results are put in the supplementary.}

\subsection{Problem Definition and Motivation}
\label{sec:motivation}
Given the exceptional perception ability of the pre-trained ViT and the instruct-following power of pre-trained LLM, we hope to combine the benefits of both worlds. Existing methods often use a generative loss (\ie captioning loss) to train an adapter module to align the vision-language modalities, which may not align vision-language effectively. To verify this, we visualize the text-token to image-path similarities of the aligned visual and text features and the results are shown in Figure~\ref{fig:motivation}. We find that the aligned visual features usually have \textit{large similarities} to all visual concepts, which may not serve as a strong condition to derive text response. In this sense, how to align the visual and language modalities effectively remains an open question.

\begin{figure*}[!t]
    \centering
    \includegraphics[width=1.0\linewidth]{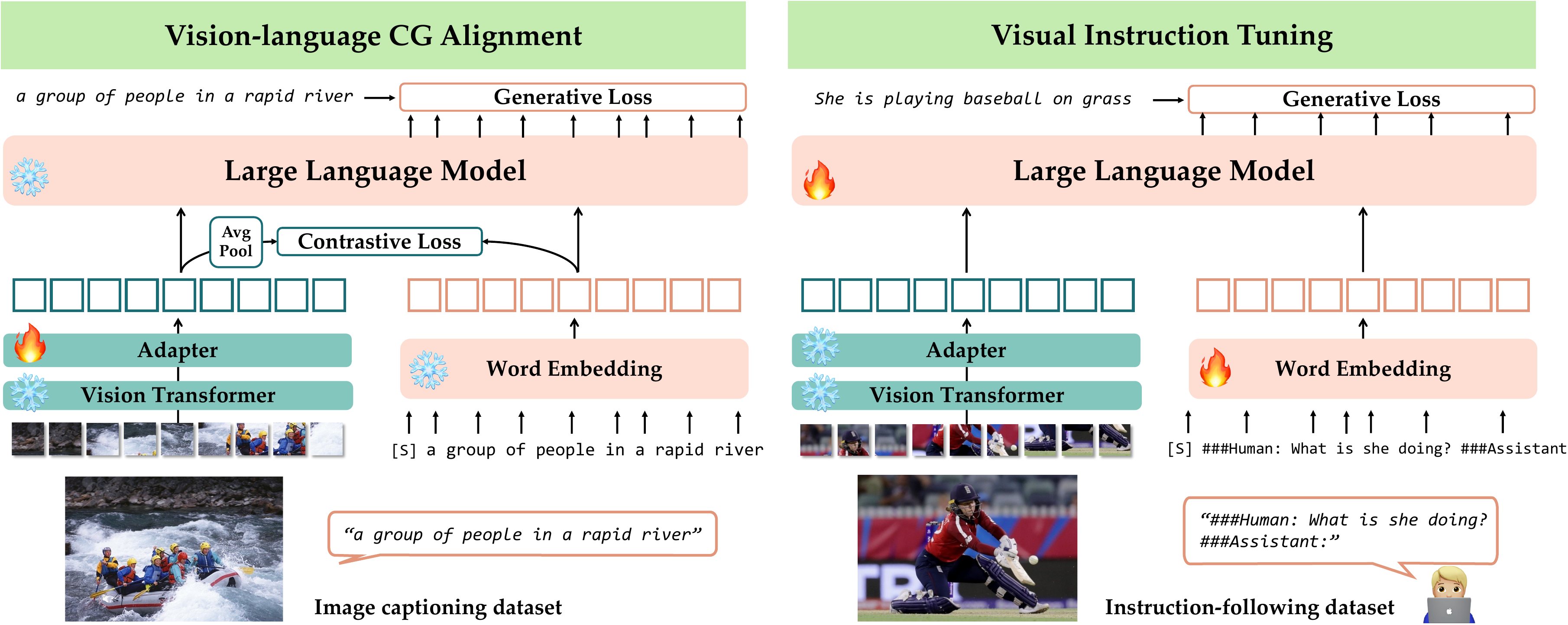}
    \caption{Overall scheme of the proposed \shortname. To endow the pre-trained large language model with visual-language instruction-following ability, our approach consists of two stages. At the vision-language CG alignment stage, we combine both contrastive and generative objectives to train a visual adapter that aligns the representation of the pre-trained vision transformer and large language model on the image captioning dataset. At the visual instruction tuning stage, we fine-tune the large language model with the instruction-following dataset to improve its instruction-following ability given visual content as the prefix.  
    % Given a pretrained ViT and a pretrained LLM, we align their features to achieve semantic harmony from two aspects. First, to resolve the semantic spaces misalignment issue, we apply a learnable linear layer on top of the ViT and a learnable \texttt{[CLS]} prompt in the LLM. Second, to resolve the semantic granularity mismatch issue, we employ learnable weighted average vectors to aggregate the semantics of different layers. Last, we conduct contrastive learning on the extracted global vision and text features to achieve vision-language alignment.
    }
    % \vspace{-10 pt}
    \label{fig:framework_vit}
    \vspace{-10pt}
\end{figure*}

\subsection{Overall Scheme of \shortname}
\label{sec:overview}
The overall scheme of our \name (\shortname) is presented in Figure~\ref{fig:framework_vit}. We transform the pre-trained ViT and LLM into a visual instruction-following model in two steps. In the first stage, we train a visual adapter with the proposed contrastive and generative alignment objectives to bridge ViT and LLM. In the second stage, following previous works~\cite{liu2023visual,liu2023improved}, we use a visual instruction-following dataset to tune the whole model except the ViT to endow the model with visual instruction-following ability.

\invisiblesection{Notations} Formally, given an image data $\bI \in \mmR^{3 \times H \times W}$, we patchify it into a visual patch sequence $\bU = \left[ u_i \right]_{i=1}^{N}$, where $N = \frac{H}{U} \times \frac{W}{U}$ and $U$ is the patch size. With a pre-trained ViT model $f(\cdot)$, we extract patch-wise features $\bV = f(\bU)$, where $\bV \in \mmR^{N \times d_v}$ and $d_v$ is the feature dimension of the ViT. Moreover, given a tokenized text data $\bX = \left[x_j \right]_{j=1}^{M} \in \mmR^{M}$, we feed it into a word embedding layer to get a sequence of text embeddings $\bE = \left[ \be_j \right]_{j=1}^M \in \mmR^{M \times d}$. We denote the LLM as $g(\cdot)$. Note that the feature dimensions of the ViT and LLM typically do not match. A projector $h(\cdot;\theta)$ is used to map the patch features into the projected features $\bZ = h(\bV;\theta)$, where $\bZ \in \mmR^{N \times d}$ and $\theta$ is the learnable parameters.

\subsection{Vision-language CG Alignment}
\label{sec:pretrain}
\invisiblesection{Vision-language generative loss} As the semantic spaces of unimodal vision and language models do not align, we need to align their representation before vision-language information propagation. To this end, existing methods often align the ViT and LLM with image captioning data \ie~paired image-text data as follows. Given the visual features and the word embeddings of prefix sequence $x_{<j}$ as inputs, the LLM predicts the probability of the next word $x_j$ as follows:
\begin{equation}
    \label{eqn:next_token_prob}
    p(x_j | u_{1:N}, x_{<j}) = g\left([\bz_i]_{i=1}^N \oplus [\be_k]_{k=1}^{j-1} \right),
\end{equation}
where $\oplus$ is the operation that concatenates visual features and word embeddings along the sequence dimension. Then, the vision-language alignment is achieved by optimizing the generative next-word prediction loss as follows:
\begin{equation}
    \mL_\text{align}^\text{gen} = - \frac{1}{M} \sum_{j=1}^M \log p(x_j | u_{1:N}, x_{<j}).
\end{equation}
As we discussed above, simply optimizing the generative alignment objective leads to weak alignment issues. Therefore, we want to align the visual features and the word embeddings as well. The intuition is simple: if the visual features can be aligned to the input word embedding, the visual features can be understood by the LLM effectively. 

\begin{table*}[t]
\small
\centering
\resizebox{\textwidth}{!}{
\begin{tabular}{@{}lrc|ccccc@{}}
\toprule
Method                                 & \begin{tabular}[c]{@{}c@{}}\#Instruction\\ Tuning Data\end{tabular} & LLM & TextVQA & VQAv2 & GQA  & \begin{tabular}[c]{@{}c@{}}SciQA-IMG\\ (0-shot)\end{tabular} & \begin{tabular}[c]{@{}c@{}}POPE\\ (0-shot)\end{tabular} \\ \midrule
Flamingo-9B~\cite{alayrac2022flamingo}  & -         & Chinchilla-7B  & -             & 51.8  & -               & -             & -                                                      \\
Flamingo-80B~\cite{alayrac2022flamingo} & -         & Chinchilla-70B & -             & 56.3  & 42.4            & -             & -                                                      \\
BLIP2~\cite{li2023blip}                 & -         & FlanT5$_{\rm XXL}$-11B     & 32.3          & 65.0  & 45.9            & 61.0          & 85.3                                                   \\
InstructBLIP~\cite{dai2023instructblip} & 1.2M      & Vicuna-13B     & 49.5          & -     & 50.7            & 63.1          & 78.9                                                   \\
Qwen-VL-Chat~\cite{bai2023qwen}         & 50M       & Qwen-7B        & 57.5          & 78.2  & 61.5            & 68.2          & -                                                      \\
Shikra~\cite{chen2023shikra}            & 5.5M      & Vicuna-13B & -             & 77.4  & -               & -             & -                                                      \\
% CogVLM-generalist~\cite{wang2023cogvlm} & 500K      & Vicuna-13B     &               & 83.4  & -               & 61.4$\dagger$ & -                                                      \\ \midrule
CogVLM~\cite{wang2023cogvlm} & 500K     & Vicuna-13B     &  -             & 83.4  & -               & ~~61.4$^\dag$ & -                                                      \\ \midrule
LLaVA-1.5~\cite{liu2023improved}        & 66K       & Vicuna-13B     & 47.1          & 65.5  & 40.7            & 62.4          & 79.5                                                   \\ 
\shortname (Ours)                              & 66K       & Vicuna-13B     & ~~~~~~~\bestimprove{53.7}{6.6} & ~~~~~~~\bestimprove{73.1}{7.6}           & ~~~~~~~~~\bestimprove{54.0}{13.3} & ~~~~~~~~\bestimprove{67.4}{5.0}             & ~~~~~~~\bestimprove{85.4}{5.9}                                          \\ \midrule
LLaVA-1.5~\cite{liu2023improved}        & 6K        & Vicuna-13B     & 43.9          & 62.1  & 37.9          & 55.7          & 77.8                                                   \\ 
\shortname (Ours)                             & 6K        & Vicuna-13B     & ~~~~~~~\bestimprove{46.6}{2.7} & ~~~~~~~\bestimprove{65.4}{3.3}  & ~~~~~~~~\bestimprove{43.7}{5.8} & ~~~~~~~~\bestimprove{61.1}{5.4} & ~~~~~~~\bestimprove{80.0}{2.2}                                          \\ \bottomrule
\end{tabular}}
% \vspace{-5pt}
\caption{Comparison with state-of-the-art visual instruction tuning methods on general VQA benchmarks. $\dag$ denotes our implementation.}
% \sxy{$\ddag$ denotes we report the results reproduced based on our re-sampled instruction tunning data subset.}}
\label{tab:sota-vq}
\end{table*}

\invisiblesection{Vision-language contrastive loss} Different from image-level and text-level representation contrastive learning like CLIP~\cite{radford2021learning}. Three difficulties lie in the alignment between ViT and LLM. \textbf{First}, compared to the image-level features of CLIP, the visual features in a vision-language framework are often patch-level to provide dense vision signals. \textbf{Second}, compared to single text features of CLIP, there is usually more than one embedding of a text caption. \textbf{Third}, there is \textit{no explicit grounding} of the visual patch and text token in the standard image captioning dataset. To address the above issues, we propose to maximize the averaged similarities between the pooled visual features and embeddings of the corresponding caption. To be specific, we first pool the image features into a global descriptor $\hat{\bZ} = \frac{1}{N}\sum_{n=1}^{N} \bZ_n$. Given a batch of image descriptors $\left[ \hat{\bZ}^1, \hat{\bZ}^2, \cdots, \hat{\bZ}^B \right] \in \mmR^{B\times d}$ and the batch word embeddings $\left[ \bE^1, \bE^2, \cdots, \bE^B \right] \in \mmR^{B \times M \times d}$ of the corresponding caption, we aim to maximize the the similarity between the paired image and text. To be specific, we compute the similarity between every image descriptor $\hat{\bZ}^i$ and text embeddings $\bE^j$ as follows:
\begin{equation}
    \label{eqn:img_token_sim}
    s^{i,j} = \frac{1}{M} \sum_{m=1}^M \epsilon\left( \hat{\bZ}^i, \bE^j_m \right),
\end{equation}
where $\epsilon(\bu,\bv) = \tau \cdot \frac{\bu^T\bv}{\lVert \bu \rVert_2 \lVert \bv \rVert_2}$ denotes scaled cosine similarity function and $\tau$ is a learnable scalar. Then, we derive our contrastive alignment loss as follows:
\begin{equation}
    \mL_\text{align}^\text{con} = \frac{1}{B} \sum_{b=1}^B \frac{\exp{\left(s^{b,b} \right)}}{\exp{\left( s^{b,b} \right) + \sum_{b'}^{b' \neq b} \exp{\left( s^{b,b'} \right)}}},
\end{equation}
where we maximize the \textit{averaged} similarity between the image descriptor and the corresponding word embeddings. 

\invisiblesection{Contrastive and generative alignment loss} Although the contrastive loss is optimized to align vision-language directly, the nature of \textit{averaged similarity} is prone to focus on the dominant visual concepts while ignoring the subtle ones. To alleviate this issue, we propose to combine both generative and contrastive loss. The intuition is that the subtle visual features must be attended to in order to correctly generate dense captions. In this sense, our contrastive and generative (CG) alignment loss is computed as follows:
\begin{equation}
    \label{eqn:alignment_loss}
    \mL_\text{align}^\text{CG} = \mL_\text{align}^\text{gen} + \alpha \mL_\text{align}^\text{con},
\end{equation}
where $\alpha$ is a hyper-parameter that controls the optimization strength of the contrastive loss. In the vision-language alignment stage, we minimize the CG alignment loss to learn the parameters of the projector.

\subsection{Visual Instruction Tuning}
\label{sec:finetune}
With the aligned representation, the visual instruction tuning stage aims to fine-tune the LLM to better follow vision-language instructions. Similar to the generative alignment process, visual instruction tuning stages also use the paired image-text data except that the text data consists of a dialogue between a human and an assistant (see Figure~\ref{fig:framework_vit}). Given the patchified input image $u_{1:N}$, the word embeddings of the text query $x_{1:Q}^q$ and the generated prefix of the text response $x_{<j}^r$, the generative loss \wrt the text response $x_{1:M}^r$ is computed as follow:
\begin{equation}
    \mL_\text{tune} = - \frac{1}{M} \sum_{j=1}^{M} \log p(x^r_j|u_{1:N}, x^q_{1:Q}, x_{<j}^r),
\end{equation}
where $p(x^r_j|u_{1:N}, x^q_{1:Q}, x_{<j}^r)$ is the probability of the next word response that is computed in the same way as in Eqn.~(\ref{eqn:next_token_prob}), given vision-language query and preceding response. Note that the text data usually has a dialogue of multiple turns. We ignore these details here for simplicity.

\section{Experiments}

\invisiblesection{Pre-training and fine-tuning datasets} 
We construct datasets for two stages: vision-language pre-training and visual instruction tuning, as done in previous works~\cite{liu2023visual,liu2023improved,dai2023instructblip,bai2023qwen}. For the pre-training stage, we leverage the LCS-558K dataset, a subset of about 558K image-text pairs from LAION-CC-SBU~\cite{schuhmann2021laion}. 
For the visual instruction tuning stage, we randomly shuffle the 665K instruction-following datasets created by LLaVA 1.5~\cite{liu2023improved} and preserve only 1\% and 10\% data to validate the model's performance under limited curated instruction-following data.\footnote{More details on the sub-sampled datasets are put in the supplementary.}
% For the visual instruction tuning stage, we fine-tune the language model on the 665K mixture of multi-task instruction-following datasets created by LLaVA 1.5~\cite{liu2023improved}. 
% For experiments on efficient instruction settings, we randomly shuffle the 665K instruction-following datasets and preserve only 1\% and 10\% data, resulting in two additional subsets with 6K and 66K data.\footnote{More details on the sub-sampled datasets are put in the supplementary.}
% We evaluate our \shortname on ScienceQA and COCO Caption benchmarks

\invisiblesection{Implementation details} We use the vision transformer of CLIP ViT-L/14~\cite{alec2021clip} and Vicuna 13B~\cite{chiang2023vicuna} to instantiate our pre-trained ViT and LLM, respectively. For text-only questions, a blank image is inputted as the placeholder into the vision transformer. During the vision-language pre-training stage, we only learn the visual adapter by freezing the parameters of pre-trained ViT and LLM. For visual instruction tuning, we only freeze the pre-trained ViT while tuning all other modules. Following the common practice of previous works~\cite{liu2023visual,liu2023improved}, the learning rate for pre-training and fine-tuning is $1\text{e}^{-3}$ and $2\text{e}^{-5}$, respectively. We use a cosine learning rate scheduler with a 0.03 warmup ratio. For both pre-training and fine-tuning, we only run through the dataset for 1 epoch. We set the penultimate visual patch features of ViT as the input for the visual adapter. The default architecture for the visual adapter is an MLP module with two linear layers and GELU~\cite{hendrycks2016gaussian} activation function.
% The batch size for pre-training and fine-tuning are set to 32 and 16 per GPU, respectively. 
The optimization strength of contrastive loss $\alpha$ is set to 1.0 default. For the LLaVA baseline and our \shortname, we fine-tune them on 66K instruction tuning data by default. All experiments are conducted on an 8$\times$A800 machine.\footnote{More implementation details can be found in the supplementary.}

\subsection{Comparison with State-of-the-arts}
To investigate the visual perception, visual-language understanding and reasoning ability of the proposed \shortname, we conduct extensive experiments on three kinds of benchmarks, namely, image captioning, visual question answering, and instruction-following.

% We evaluate our \shortname on various multi-modal tasks to investigate the visual perception and visual-language understanding ability. Specifically, we perform three types of multi-modal tasks: image captioning, visual question answering, and instruction-following.

\begin{table}[t]
\centering
\small
\resizebox{0.45\textwidth}{!}{
\begin{tabular}{@{}lr|cc@{}}
\toprule
Method                                 & \begin{tabular}[c]{@{}c@{}}\#Instruction\\ Tuning Data\end{tabular} & MME    & SEED \\ \midrule
BLIP2~\cite{li2023blip}                 & -    & 1293.8 & 46.4 \\
MiniGPT4~\cite{zhu2023minigpt}          & -    & 581.7  & 47.4 \\
InstructBLIP~\cite{dai2023instructblip} & 1.2M & 1212.8 & 58.8 \\
% Shikra~\cite{chen2023shikra}            & \sxy{?}  & -      & -    \\
LLaMA-AdapterV2~\cite{gao2023llama}     & 50M  & 972.7  & 35.2 \\
LAVIN~\cite{luo2023cheap}               & 210K & 963.6  & -    \\
LLaVA~\cite{liu2023llava}               & 158K & 502.8  & 37.0 \\
mPLUG-Owl~\cite{ye2023mplug}            & 392K & 967.3  & 37.9 \\
Qwen-VL-Chat~\cite{bai2023qwen}         & 50M  & 1487.5 & 65.4 \\ \midrule
LLaVA-1.5~\cite{liu2023improved}        & 66K  & 1061.8 & 51.1 \\ 
\shortname (Ours)                       & 66K  & \textbf{1318.2} & \textbf{60.9} \\ \midrule
LLaVA-1.5~\cite{liu2023improved}        & 6K   & 1051.8 & 45.3 \\ 
\shortname (Ours)                       & 6K   & \textbf{1143.2} & \textbf{51.4} \\ \bottomrule
% LLaVA-1.5~\cite{liu2023improved}        & Vicuna-13B & 1531.3 & 61.6 & 67.7    & 63.4        \\ \midrule
% \shortname (Ours)                       & Vicuna-13B &        &      &         &             \\ \bottomrule
\end{tabular}}
% \vspace{-5pt}
\caption{Comparisons with state-of-the-art visual instruction tuning methods on human instruction-following benchmarks.}
\label{tab:sota-ins}
\end{table}

\begin{table}[t]
\small
\centering
\resizebox{0.47\textwidth}{!}{
\begin{tabular}{@{}ll|ccc@{}}
\toprule
\multirow{2}{*}{Method} & \multirow{2}{*}{\begin{tabular}[c]{@{}l@{}}Alignment\\ Objective\end{tabular}} & \multicolumn{2}{c}{ScienceQA} & \multirow{2}{*}{POPE} \\ \cmidrule(lr){3-4}
                        &                                                                                & Text        & Image       &                       \\ \midrule
LLaVA~\cite{liu2023improved}  & Gen.                                                                           & 73.7        & 62.4        & 81.3                  \\
\multirow{2}{*}{\shortname}    & Con.                                                                           & 72.2        & 59.7        & 80.9                  \\
                        & Con.+ Gen.   & ~~~~~~~\bestimprove{77.0}{3.3}                     & ~~~~~~~\bestimprove{67.4}{5.0}                      & ~~~~~~~\bestimprove{85.4}{4.1}                  \\ \bottomrule
\end{tabular}}
% \vspace{-5pt}
\caption{Ablation studies on different alignment objectives. Both LLaVA and our CG-VLM are trained on the 10\% subset of instruction-tuning data. Gen. and Con. are short for Generative and Contrastive, respectively.}
\label{tab:ablation-align}
\end{table}

\invisiblesection{Results on visual question answering} We first evaluate our model on the object probing dataset (POPE~\cite{li2023pope}). In this task, the vision-language model is asked to answer a series of questions about the existence of specific objects in the image. From the first column in Table~\ref{tab:sota-vq}, we found that our \shortname model produced more accurate probing results since it is well-trained to distinguish objects through image-text paired contrastive learning. Using only 6K instruction tuning data, our \shortname model surpasses InstructBLIP trained on a much larger data scale (80.0 \vs 78.9), which shows the effectiveness of contrastive alignment for accurate perception and less object hallucination.
% preventing the model from the object hallucination problem.
% Additionally, to verify whether our VLM model suffers from the object hallucination problem, we further adopt a polling-based query benchmark POPE~\cite{li2023pope} to examine our \shortname model.

We further test our model on another two variants of the visual question-answering benchmarks, including text-oriented VQA and general VQA benchmarks. For text-oriented VQA benchmarks, we adopt TextVQA~\cite{singh2019textvqa} which requires the VLM model to read the text within images and reason the answers. For general VQA, we utilize three benchmarks various from VQAv2~\cite{goyal2017vqav2}, GQA~\cite{hudson2019gqa} to ScienceQA~\cite{lu2022sqa} to examine visual perception and reasoning ability. We report direct comparison results with a generative pre-training baseline (LLaVA-1.5) under two different data fractions, 6K and 66K. Overall performance on all VQA tasks is reported in Table~\ref{tab:sota-vq}. As the results show, our \shortname significantly improves the model's ability among all VQA tasks. It is worth noting that on the GQA benchmark, which consists of complex compositional questions, our \shortname achieves 54.0\% overall accuracy when finetuned on 66K instruction data, compared to 40.7\% of the baseline model. We attribute it to our alignment objective. By explicitly reducing the modality gap between visual and language inputs during the pre-training stage, it enables accurate perceiving and reasoning of objects in the image.

\invisiblesection{Results on instruction-following} For modern VLM models, the ability of instruction-following is crucial and of essential concern in real-world scenarios where diverse commands will be encountered. To investigate our \shortname model's performance in this scenario, we conduct evaluations on the MME~\cite{fu2023mme} and SEED~\cite{li2023seed} benchmarks. We report the perception score for the MME benchmark. For the SEED benchmark, we report accuracy on the image set. Results are shown in Table~\ref{tab:sota-ins}.
Despite extremely little training data, our \shortname still yields better results on both MME and SEED benchmarks, compared with previous methods trained on much larger datasets (\ie InstructBLIP and LLaMA-AdapterV2). We also achieved competitive performance against the data-hungry method Qwen-VL-Chat (1318.2 \vs 1487.5), which is trained on 1.4B image-text pairs and 50M multi-task instruction data. These results demonstrate that \shortname is able to adapt rapidly to a small amount of instruction tuning data while preserving strong generalization ability to unseen instructions.
% demonstrates competitive performance against methods that trained on heavy instruction tuning datasets (\ie, Qwen-VL-Chat that trained on 50M data, ours 66K) 

\subsection{Ablation Analysis on \shortname}
% \vspace{-5pt}
We conduct ablation studies to investigate the effectiveness of different alignment objectives: generative-only (LLaVA~\cite{liu2023improved}) and using both contrastive and generative objectives (our \shortname). The POPE~\cite{li2023pope} and ScienceQA~\cite{lu2022sqa} benchmarks are selected to evaluate the abilities of visual perception, visual-language reasoning and language-only reasoning. The results are shown in Table~\ref{tab:ablation-align}. Compared to the generative objective, using the contrastive objective only leads to performance degradation on all metrics, showing that discarding the generative objective totally to learn vision-language alignment is not feasible, especially when no explicit grounding information is provided for image-patch and text token. Instead, our \shortname combines both contrastive and generative objectives, which not only directly aligns the dominant visual concepts effectively, but also aligns the non-dominant ones supervised by the generative loss. From Table~\ref{tab:ablation-align}, we observe remarkable improvements in both datasets after pre-training using our proposed contrastive alignment objective. Specifically, for visual perception ability, \shortname improves the results on POPE dataset by 4.1\%. For vision-language understanding, we observe improvements in both visual (67.4 \vs 62.4) and text (77.0 \vs 73.7) modalities on the ScienceQA dataset. These findings support the superiority of using both contrastive and generative objectives other than generative only.

% \invisiblesection{Effectiveness of soft prompt tuning}

% \subsection{Further Analysis on \shortname}
% To gain deeper insights into our \shortname, we further conduct a series of ablation studies of the detailed design \sxy{during the instruction process}.

% \begin{table}
% \centering
% \small
% \begin{tabular}{@{}cc|cc@{}}
% \toprule
% Align method                & Tune proj.        & SciQA-IMG & POPE   \\ \midrule
% \multirow{2}{*}{LLaVA}      & \tiny\Checkmark   &  62.4     & 81.3  \\
%                             & \tiny\XSolidBrush &           &      \\ \midrule
% \multirow{2}{*}{\shortname (Ours)} & \tiny\Checkmark   &  66.4     & 86.1 \\
%                             & \tiny\XSolidBrush &  67.4     & 85.4  \\ \bottomrule
% \end{tabular}
% \caption{Ablation studies on projector learning strategy.}
% \label{tab:ablation-proj-tune}
% \end{table}

\subsection{Further Analysis on \shortname}
% \vspace{-5pt}
\invisiblesection{Effect of $\alpha$ in Eqn.~(\ref{eqn:alignment_loss})}
Recall that $\alpha$ is used to control the optimization strength of contrastive loss in the vision-language alignment stage. We investigate its effect on the visual instruction-tuning performance of ScienceQA and POPE by setting $\alpha$ to 0.5, 1.0 and 2.0. From the results in Table~\ref{tab:ablation-alpha}, we observe our \shortname is relatively robust to the choices of $\alpha$ when $\alpha \geq 1.0$ and the best results are achieved when $\alpha=1.0$. When decreasing $\alpha$ to $0.5$, the performance on both ScienceQA and POPE drops substantially, indicating that strong optimization strength on the contrastive loss is important to achieve better performance.

\invisiblesection{Effect of projector design} Since the projector is optimized to minimize both generative and contrastive objectives, we are curious about its capacity. We investigate the capacity of the projector by instantiating it into an MLP layer or a linear layer and the results are shown in Table~\ref{tab:ablation-proj-design}. We find that using the MLP projector achieves the best results, indicating that an adapter with strong learning capacity is able to facilitate the vision-language alignment process.

\begin{table}[t]
\centering
\small
\begin{tabular}{r|ccc}
\toprule
$\alpha$  &  0.5  & 1.0  & 2.0 \\ \midrule
SciQA-IMG &  64.5 &  \textbf{67.4} & 67.1    \\
POPE      &  84.4 &  \textbf{85.4} & \textbf{85.4}    \\ \bottomrule
\end{tabular}
\vspace{-5pt}
\caption{Ablation studies on $\alpha$ in Eqn.~(\ref{eqn:alignment_loss}).}
\label{tab:ablation-alpha}
\vspace{-5pt}
\end{table}

\begin{table}[h]
\centering
\small
\begin{tabular}{r|cc}
\toprule
Projector Type & MLP & Linear \\ \midrule
SciQA-IMG      &  \textbf{67.4}     &  64.0 \\
POPE           &  \textbf{85.4}     &  81.0  \\ \bottomrule
\end{tabular}
\vspace{-5pt}
\caption{Ablation studies on the projector type.}
\label{tab:ablation-proj-design}
\vspace{-5pt}
\end{table}

\begin{table}[h]
\centering
\small
\begin{tabular}{r|ccc}
\toprule
Ratio  & 1\% & 10\% & 100\% \\ \midrule
SciQA-IMG &  47.6 &  62.1 & 67.4    \\
POPE &  49.2 &  67.7 & 85.4    \\ \bottomrule
\end{tabular}
\vspace{-5pt}
\caption{Ablation studies on different amounts of pre-trained data.}
\label{tab:ablation-pretrain-data}
\vspace{-5pt}
\end{table}

\invisiblesection{Effect of pre-trained data amount} Although pre-trained data is much easier to obtain than the instruction tuning data, we are still curious about how the amount of pre-trained data affects the visual instruction-following performance. We conduct experiments by setting the amount of pre-trained data to 1\%, 10\% and 100\%. From the results in Table~\ref{tab:ablation-pretrain-data}, our \shortname achieves competitive results with only 10\% pre-trained data on the SciencQA image set, showing that \shortname has great potential for efficient vision-language alignment. 
% We also observed gradual improvement as the pre-training data scale rises. Since it is much easier to collect image-text pairs for pre-training, our approach holds significant promise as the scale of pre-training data expands.

% \invisiblesection{Logit scale} xxx

% \invisiblesection{Prompt length} xxx

% \invisiblesection{Text select layer}

% \invisiblesection{Original Text ability} xxx

% \invisiblesection{Sort Prompt Position} xxx

% \invisiblesection{Effect of Pretrain Batch Size} xxx

\subsection{Qualitative Results of \shortname} 
\begin{figure*}[!th]
    \centering
    \includegraphics[width=0.98\linewidth]{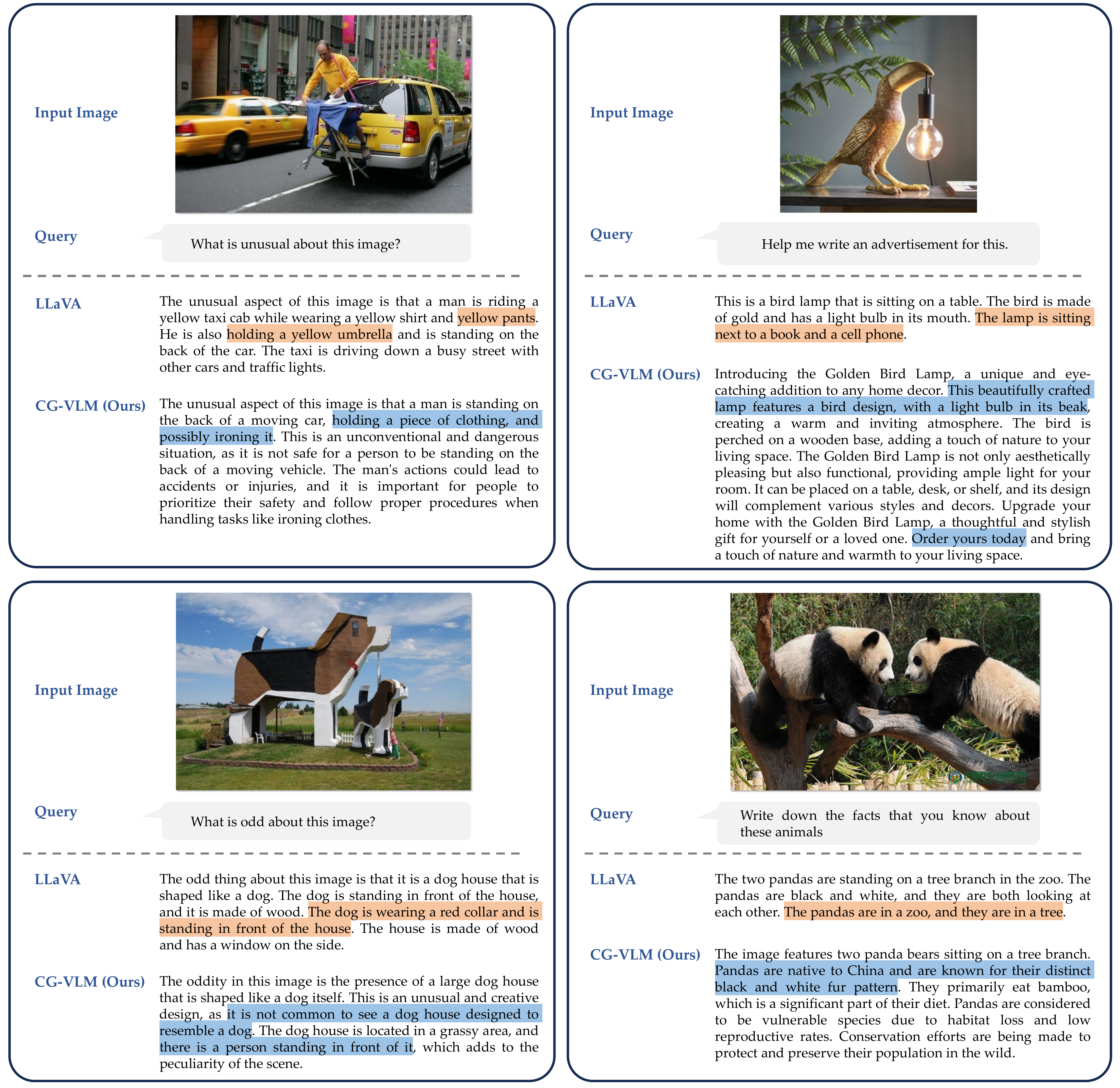}
    \vspace{-9pt}
    \caption{Qualitative comparisons of LLaVA and our~\shortname on general-purpose visual-language queries. We highlight the results of the favorable parts and the unfavorable parts with \colorbox{favorable}{blue shading} and \colorbox{unfavorable}{orange shading}, respectively.\protect\footnotemark
    }
    \label{fig:visualization}
    \vspace{-15pt}
\end{figure*}

\vspace{-5pt}
To intuitively understand our \shortname's ability of visual perception, vision-language understanding and visual instruction-following, we provide visual comparison results between LLaVA 1.5~\cite{liu2023improved} and our \shortname in Figure~\ref{fig:visualization}. Note that both LLaVA 1.5 and \shortname are fine-tuned on the 10\% subset of data. We construct our test cases to evaluate the abilities of visual instruction models in three aspects.

\invisiblesection{Visual perception and reasoning ability} From the first column, we find that our \shortname accurately perceives and reasons about the visual concepts (``holding $\cdots$ clothing and possibly ironing it''), while LLaVA can not accurately capture relevant visual concepts and fails to understand the odd context in the image
% (``yellow pants'', ``holding a yellow umbrella'').
(``holding a yellow umbrella'').

\invisiblesection{Visual instruction-following ability} As shown in the top right, our \shortname understands the query and completes the required task based on rich visual details (``a light bulb in its beak'', ``Order yours today''), while LLaVA fails to follow instruction and only produces a caption-like response (``sitting next to a book and a cell phone''). 

\invisiblesection{Original LLM ability} As depicted in the bottom right, our \shortname preserves more original LLM ability and gives a detailed introduction of pandas (``native to China'', ``known for their $\cdots$''), while LLaVA can only roughly describe the image content (``in a zoon and in a tree'').

% \subsection{Comparison of data efficiency}
% As discussed in previous work~\cite{dai2023instructblip}\sxy{[TODO], the quality and amount of training data in the instruction tunning stage is crucial for VLM to master complex vision-language interaction.} We compare the visual question answering performance under different data scales in the instruction tunning stage and find out our \shortname is a data-efficient learner. \sxy{[TODO] (a chart)} We surprisingly observed that our \shortname surpasses most of the previous SOTA VLM models even though only 66K instruction-following data are available. \sxy{[TODO] (further analyses)}

\section{Conclusion}
% \vspace{-10pt}
In this work, we investigate the vision-language alignment issue when extending large language models to the vision-language instruction-following model. We find that existing methods often produce weak alignment by optimizing the generative image captioning loss only and are highly dependent on instruction-tuning data that is very hard to collect. To improve the vision-language alignment of existing methods, we propose a \name (\shortname) method that combines both contrastive and generative objectives for vision-language alignment. The contrastive objective is proposed to maximize the averaged similarities between pooled image features and text word embeddings. With the contrastive objective that aligns the vision-language directly, the generative objective is also preserved to learn the alignment of non-dominant visual concepts ignored by the averaged mechanism. Extensive experiments on various downstream tasks show that our \shortname is an efficient instruction learner. In the future, we will explore the contrastive objective given the grounding relation between the image patch and the text token.
\footnotetext{More qualitative results can be found in the supplementary.}

% \input{sec/6_todo}
% \clearpage

% \noindent\textbf{Acknowledgements.} {This work was partially supported by Key-Area Research and Development Program of Guangdong Province 2019B010155001, National Natural Science Foundation of China (NSFC) (62072190), National Natural Science Foundation of China (NSFC) 61836003 (key project), Program for Guangdong Introducing Innovative and Entrepreneurial Teams 2017ZT07X183.}

{
    \small
    \bibliographystyle{ieeenat_fullname}
    \bibliography{main}
}


% \includepdf[pages=-]{CG-VLM-supp-arxiv-202311241755.pdf}
% \includepdf[pages=-]{CG-VLM-supp-arxiv-202311282108.pdf}


\section*{Supplementary material}

# Contrastive Vision-Language Alignment Makes Efficient Instruction Learner (Supplementary Materials)

Lizhao Liu[1,2][*] Xinyu Sun[1][*] Tianhang Xiang[1] Zhuangwei Zhuang[1] Liuren Yin[3] Mingkui Tan[1,2][†]
[1]South China University of Technology  [2]PengCheng Laboratory  [3]Duke University
{selizhaoliu, csxinyusun, sexiangtianhang, z.zhuangwei}@mail.scut.edu.cn,
liuren.yin@duke.edu, mingkuitan@scut.edu.cn

We organize our supplementary materials as follows:
- In Section A, we provide more implementation details on the proposed method.
- In Section B, we provide more details on the instruction-following dataset.
- In Section C, we provide more visualization results on the vision-language alignment.
- In Section D, we provide more qualitative results on visual-language queries.
- In Section E, we provide failure case analysis to understand the limitations of the proposed method.

## A. More Implementation Details

We build our code base on the huggingface [15] framework. We use the batch size of 32 and 128 for pre-training and fine-tuning, respectively. For image preprocessing, For the text tokenizer, we set the maximum sequence length to 2,048. With 576 (*i.e.,* $\frac{336}{14}^2$) image patch features from CLIP ViT-L/14 336 px [11], the maximum input sequence length to large language model (LLM) is 2,624. Note that we treat the visual features as a prefix to LLM and do not use extra tokens to denote the start or the end of the image patch features. For pre-training, the training time is about 8 hours. For fine-tuning, the training time is about **15 minutes** and **2 hours** for **1%** and **10%** instruction-following data, respectively.

## B. More Details on Instruction-following Datasets

Based on the instruction-following dataset from LLaVA-665K [6], we randomly shuffle and sub-sample it to construct two 1% and 10% variants, LLaVA-6K and LLaVA-66K. Here, we provide detailed dataset statistics comparisons on LLaVA-6K, LLaVA-66K and LLaVA-665K. The results are shown in Table I, where all variants retain the same data size distribution.

## C. More Results on Vision-language Alignment

To demonstrate that our CG-VLM effectively aligns the vision-language modalities, we provide more qualitative results of vision-language similarity in Figure I. We have two observations. **First**, our CG-VLM is better at aligning the main visual concepts in the image such as the shirt, watch and dragon in rows 1, 2 and 3, respectively. In contrast, LLaVA aligns not only with the main visual concepts but also with irrelevant concepts even the background. **Second**, we find that the proposed CG-VLM is better at aligning the visual sub-parts such as the rose, strap, the text in rows 1, 2 and 4, respectively. We admit that, even with the contrastive alignment, our CG-VLM can not align the vision-language modalities perfectly if no grounding relation between the image patch and the text token provided. In the future, we will explore how to use more advanced models such as GLIP [5] and GLIP V2 [16] to conduct more strong vision-language alignment to improve the visual instruction-tuning models.

## D. More Qualitative Results on Visual-Language Queries

In the main paper, we provide the qualitative comparisons on the LLaVA [6] baseline and the proposed CG-VLM that are both trained with 10% of instruction-tuning data. Here, we provide more results of models trained on 1% of instruction-tuning

---

[*]Equal contribution.
[†]Corresponding author.

1

| Dataset | Size | | | Response formatting prompts |
|---|---|---|---|---|
| ShareGPT [13] | 40K | 4K | 0.4K | - |
| LLaVA [7] | 158K | 16K | 1.6K | - |
| VQAv2 [1] | 92K | 9K | 0.9K | |
| OKVQA [9] | | | | Answer the question using a single word or phrase. |
| GQA [2] | 72K | 7K | 0.7K | |
| OCRVQA [10] | 80K | 8K | 0.8K | |
| A-OKVQA [12] | 66K | 7K | 0.6K | Answer with the option's letter from the given choices directly. |
| TextCaps [14] | 22K | 2K | 0.2K | Provide a one-sentence caption for the provided image. |
| RefCOCO [3, 8] | 48K | 5K | 0.5K | Provide a short description for this region. (or) |
| VG [4] | 86K | 9K | 0.9K | Provide the bounding box coordinate of the region this sentence describes. |
| Total | 665K | 66K | 6K | |

Table I. Dataset statistics for three variants of LLaVA-1.5 instruction-following dataset: 665K, 66K and 6K, which correspond to 100%, 10% and 1% of the original LLaVA-665K dataset, respectively.

data. The qualitative comparisons are conducted to evaluate three aspects: visual-language reasoning, visual-instruction following and original language ability. We summed up CG-VLM's advantage as follows:

**Better at vision-language reasoning**. As shown in Figure II, we generally find that our CG-VLM captures the relevant visual concepts accurately and conducts convincingly reasoning to respond to the query. For example, in the top left example, trained with 10% of instruction-tuning data, our CG-VLM is able to perceive that "a man is standing on the back of a moving car, holding a piece of clothing and possibly ironing it" and understand that "it is not safe for a person to be standing on the back of a moving vehicle". Instead, LLaVA shows a poor ability at visual perception *i.e.,* "wearing yellow pants and holding a yellow umbrella" and is unable to derive a corresponding response to the query.

**Better at visual-instruction following**. As shown in Figure III, we generally find that our CG-VLM is able to effectively follow the instruction and give a favorable response to the query, even trained with only 1% of instruction tuning data. For instance, our CG-VLM is able to follow various kinds of instructions such as advertisement (selling the gold bird statue and the ultimate bathroom ), introduction (introducing a Bernese Mountain Dog) and using literary language (describing scene using literary language). Surprisingly, our CG-VLM, trained with 10% of data, names the flying bathroom from the top right image as "Shittee", showing its great creativity. In contrast, the LLaVA baseline, trained with 1% or 10% of data, merely functions as an image captioner, describing the contents of the image without following the instructions.

**Better at original LLM language ability**. As shown in Figure IV, we generally find that our CG-VLM is able to preserve more of the original LLM language ability. For example, trained with 1% or 10% of data, our CG-VLM is able to provide the background information of the panda ("native to China", "known for their distinct black and white fur pattern") and write a beautiful poem to describe the image. Note that, our CG-VLM trained with 10% of data, even constructs a poem that begins with a rhyming sentence *i.e.,* "In the golden sunset's glow, A man and his dot sit close", demonstrating the great language ability preserved from the original LLM. However, the LLaVA baseline fails to provide the background knowledge of pandas and can not write a poem as beautiful as our CG-VLM.

## E. Failure Case Analysis of the proposed CG-VLM

In this section, we provide some failure case analysis to understand the limitations of the proposed CG-VLM. We find that our CG-VLM has two main limitations. **First**, the contrastive alignment neglects some image details in some cases, though we apply the generative loss to capture the details or concepts of the image. From the left part of Figure V, we see that our CG-VLM misidentifies the sentence "Monday Just ... Monday" as "Monday, just like just about every other day", showing the contrastive objective of CG-VLM can not effectively recognize some dense text concepts. Note that this issue can be alleviated by using reference OCR tokens as inputs [6]. **Second**, we also find that both the LLaVA baseline and our CG-VLM are dominated by the text modality in some cases. For instance, in the right part of Figure V, we find that both LLaVA and CG-VLM are misled by the text content in the image *i.e.,* "I'm good.", where the cat is not good instead. In the future, we will investigate how to balance the vision-language modalities to produce reliable reasoning results.

|              | (a) Image and Caption | (b) Text-token to Image-patch Similarities |
|---|---|---|

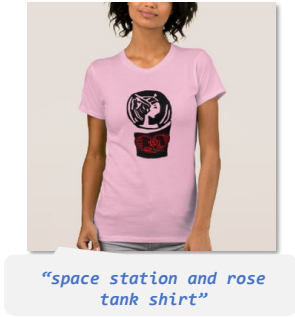

"space station and rose tank shirt"

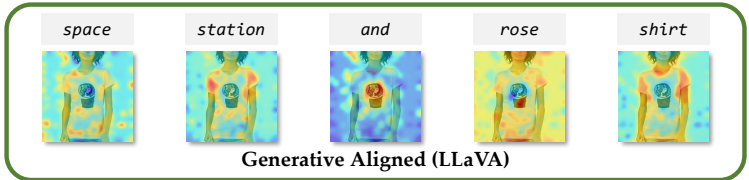
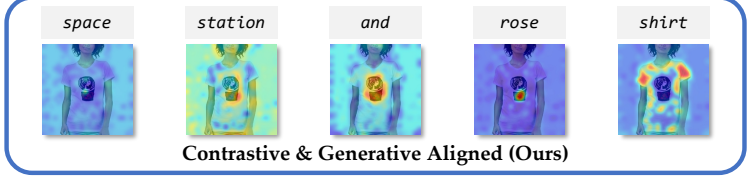

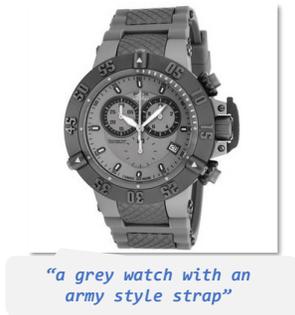

"a grey watch with an army style strap"

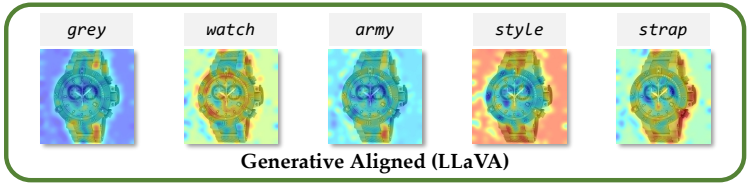
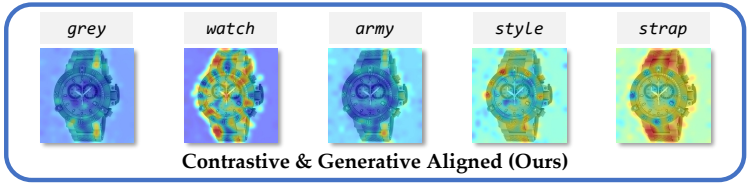

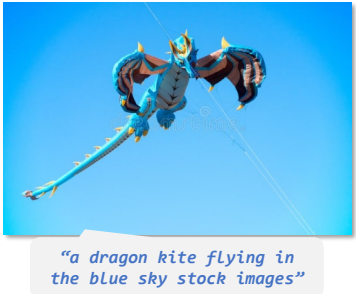

"a dragon kite flying in the blue sky stock images"

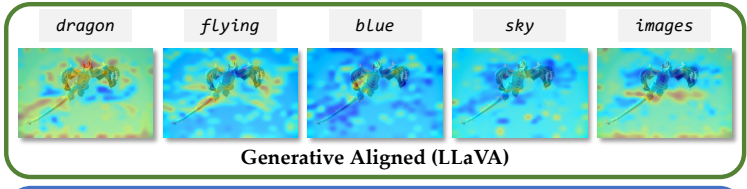
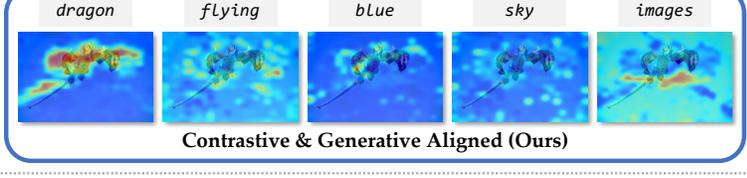

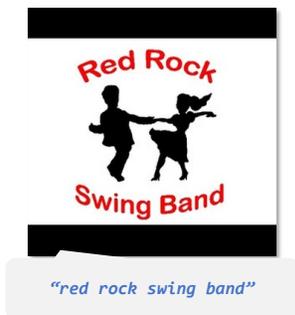

"red rock swing band"

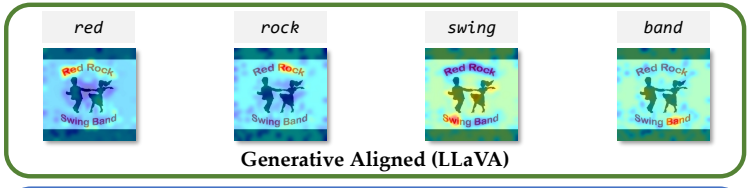
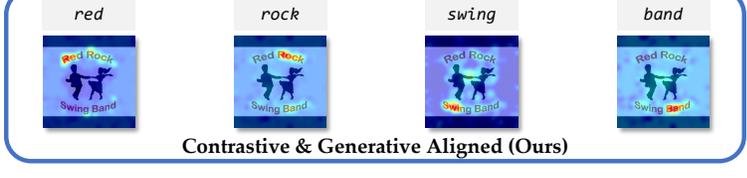

Figure I. More comparisons of image features projected by different visual adapters. With generative objective only, the projected visual features of LLaVA [6] often have large similarities to all visual concepts. In contrast, our CG-VLM uses both contrastive and generative alignment objectives to train the visual adapter, providing large similarities *only* to the corresponding concepts.

3

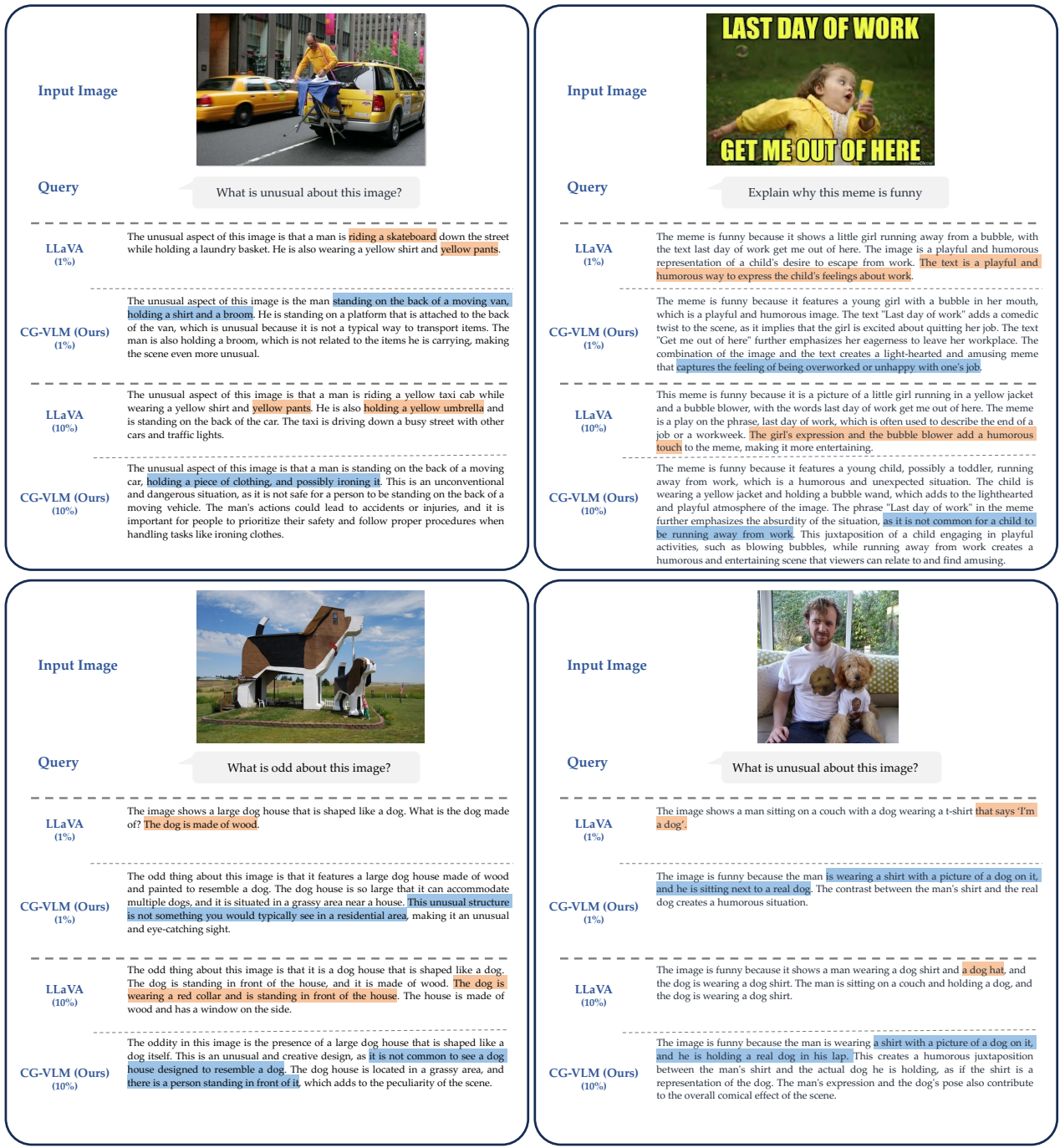

Figure II. More qualitative comparisons of LLaVA [6] and our CG-VLM on vision-language reasoning ability. We highlight the results of the favorable parts and the unfavorable parts with blue shading and orange shading, respectively.

4

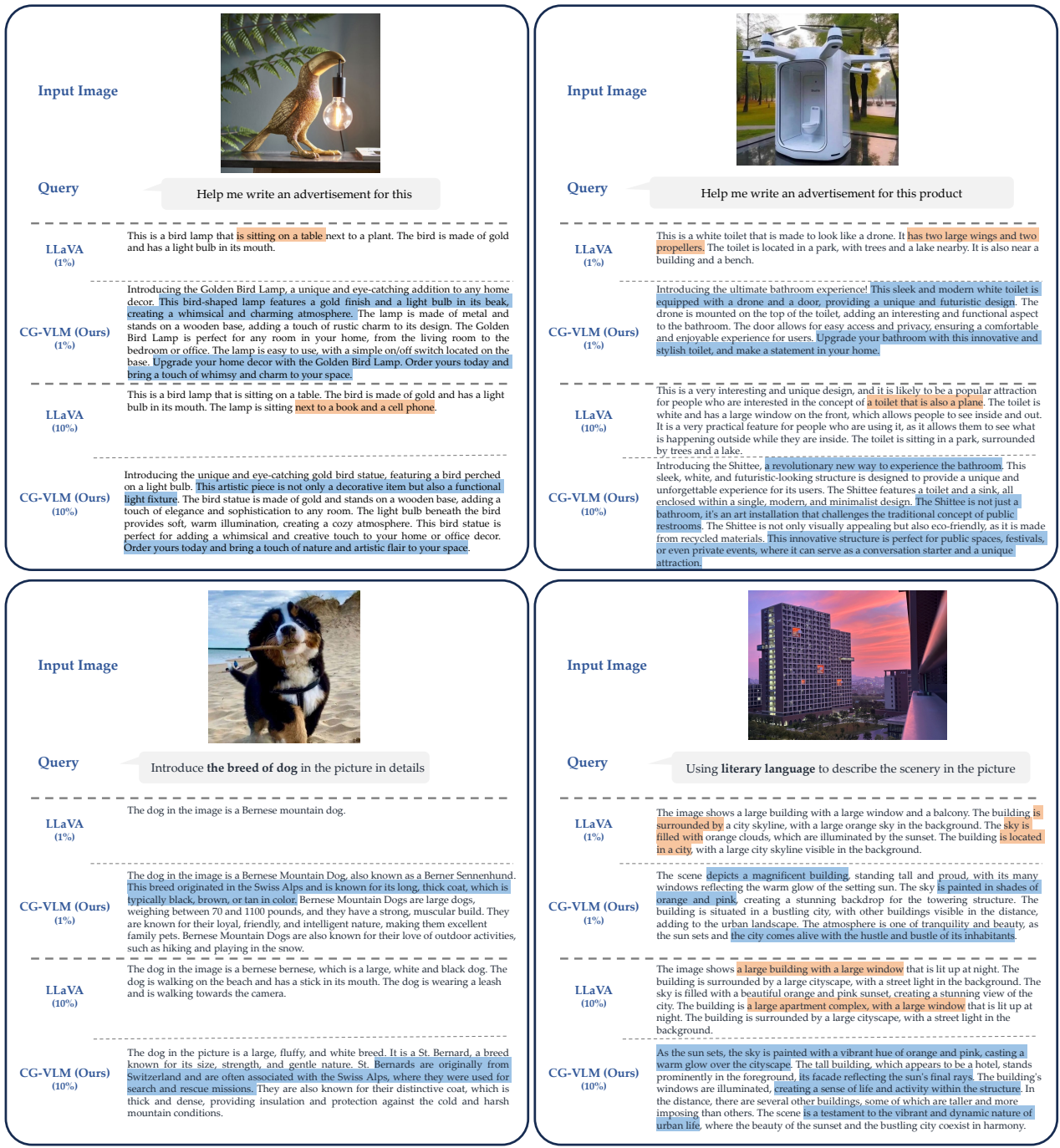

Figure III. More qualitative comparisons of LLaVA and our CG-VLM on vision-language instruction-following ability. We highlight the results of the favorable parts and the unfavorable parts with blue shading and orange shading, respectively.

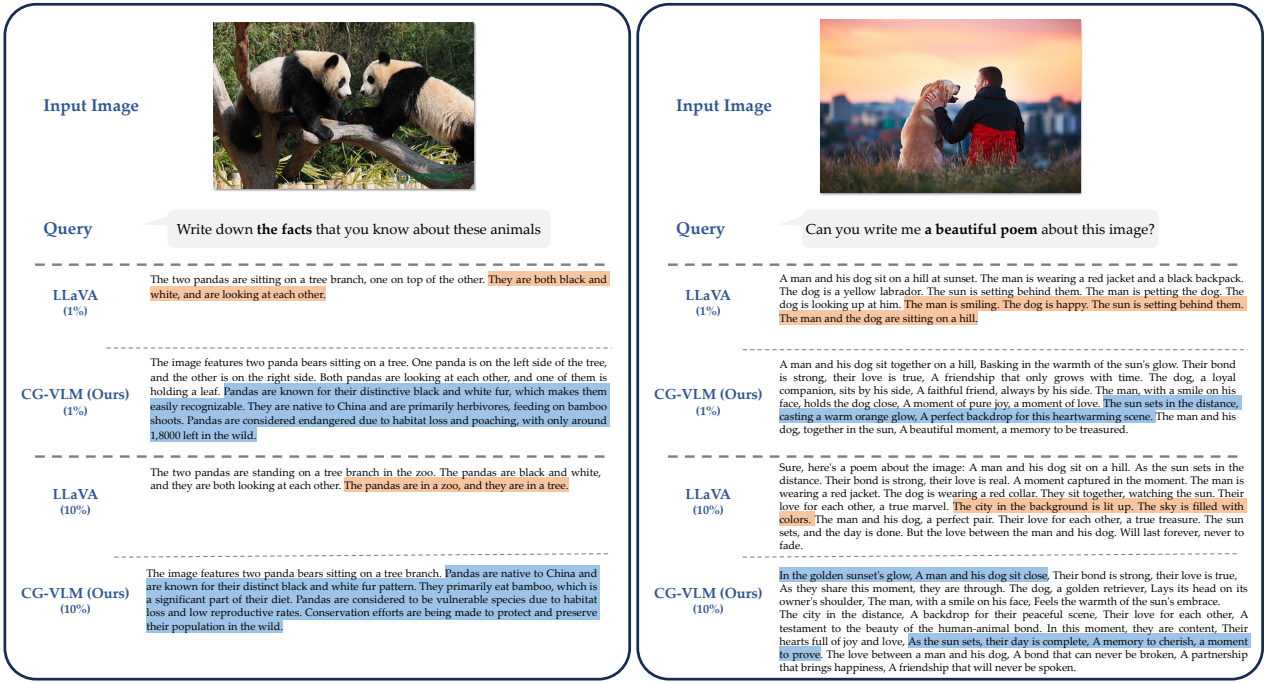

Figure IV. More qualitative comparisons of LLaVA and our CG-VLM on language ability ability. We highlight the results of the favorable parts and the unfavorable parts with blue shading and orange shading, respectively.

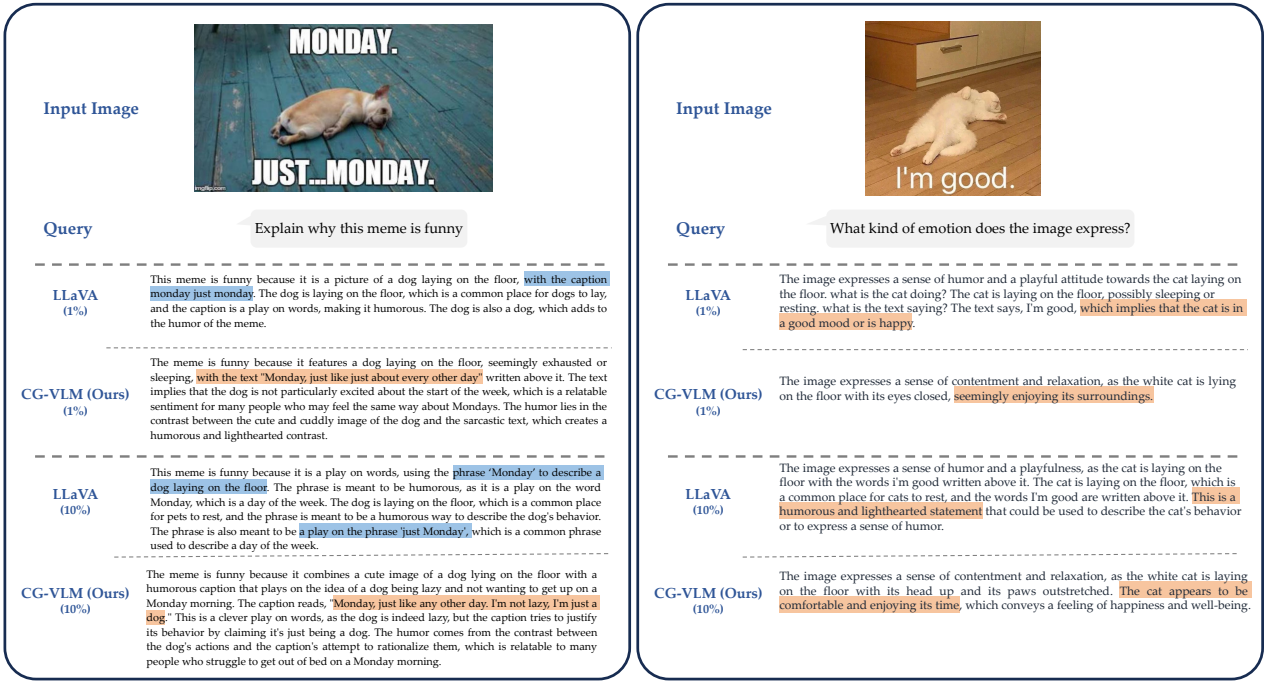

Figure V. Failure case of LLaVA and our CG-VLM, which shows that the model is misled by the dominant concept. We highlight the results of the favorable parts and the unfavorable parts with blue shading and orange shading, respectively.



\end{document}